%% file: main.tex
\documentclass{article}

\usepackage{microtype}
\usepackage{graphicx}
\usepackage{subfigure}
\usepackage{booktabs} 

\usepackage{hyperref}

\usepackage[accepted]{icml2021}

\usepackage[para]{threeparttable}
\usepackage{multirow}
\usepackage{graphicx}
\usepackage{amsmath}
\usepackage{amssymb}
\usepackage{hyperref}
\usepackage{makecell}
\usepackage{pifont}
\usepackage{xspace}
\usepackage{adjustbox}

\newcommand{\cmark}{\ding{51}}%
\newcommand{\xmark}{\ding{55}}%
\DeclareMathOperator*{\argsort}{argsort}
\newcommand{\frameworkname}{SeViT}

\icmltitlerunning{Semi-Parametric Video-Grounded Text Generation}

\begin{document}

\twocolumn[
\icmltitle{Semi-Parametric Video-Grounded Text Generation}




\begin{icmlauthorlist}
\icmlauthor{Sungdong Kim}{naver,kaist}
\icmlauthor{Jin-Hwa Kim}{naver,snu}
\icmlauthor{Jiyoung Lee}{naver}
\icmlauthor{Minjoon Seo}{kaist}

\end{icmlauthorlist}

\icmlaffiliation{naver}{NAVER AI Lab}
\icmlaffiliation{kaist}{KAIST AI}
\icmlaffiliation{snu}{SNU AIIS}

\icmlcorrespondingauthor{Minjoon Seo}{minjoon@kaist.ac.kr}


\vskip 0.3in
]



\printAffiliationsAndNotice{}

\input{mysymbol.tex}
\input{tabs/00_abstract.tex}
\input{tabs/01_introduction.tex}
\input{tabs/02_background.tex}
\input{tabs/03_method.tex}

\input{tabs/04_experimental_setup.tex}
\input{tabs/05_experimental_result.tex}
\input{tabs/06_conclusion.tex}

\bibliography{references}
\bibliographystyle{icml2021}

\input{tabs/08_appendix.tex}

\end{document}

%% file: mysymbol.tex
\def\eg{\emph{e.g}.\xspace}
\def\ie{\emph{i.e}.\xspace}

%% file: tabs/00_abstract.tex
\begin{abstract}
Efficient video-language modeling should consider the computational cost because of a large, sometimes intractable, number of video frames. Parametric approaches such as the attention mechanism may not be ideal since its computational cost quadratically increases as the video length increases. Rather, previous studies have relied on offline feature extraction or frame sampling to represent the video efficiently, focusing on cross-modal modeling in short video clips. In this paper, we propose a semi-parametric video-grounded text generation model, \frameworkname{}, a novel perspective on scalable video-language modeling toward long untrimmed videos. Treating a video as an external data store, \frameworkname{} includes a non-parametric frame retriever to select a few query-relevant frames from the data store for a given query and a parametric generator to effectively aggregate the frames with the query via late fusion methods. Experimental results demonstrate our method has a significant advantage in longer videos and causal video understanding.
Moreover, our model achieves the new state of the art on four video-language datasets, iVQA (+4.8), Next-QA (+6.9), and Activitynet-QA (+4.8) in accuracy, and MSRVTT-Caption (+3.6) in CIDEr.

\end{abstract}

%% file: tabs/01_introduction.tex
\section{Introduction}

Recently, there has been the impressive success of vision-language models~\cite{lu2019vilbert, radford2021learning, li2022blip, alayrac2022flamingo} demonstrating a remarkable transferability on video-language tasks including video retrieval, video captioning, and video question answering (Video QA). Considering video input's spatio-temporal aspects, it is often more challenging to process than other modalities, especially combining video and language modalities. Moreover, modeling video information requires heavy computations since it comprises lengthy image sequences (frames). Conventionally, many video-language works have relied on the pre-trained vision or video encoder as an offline feature extractor to represent video densely but efficiently~\cite{sun2019videobert, li2020hero, yang2021just}.

\begin{figure}[t!] 
\centering
\includegraphics[width=0.9\columnwidth]{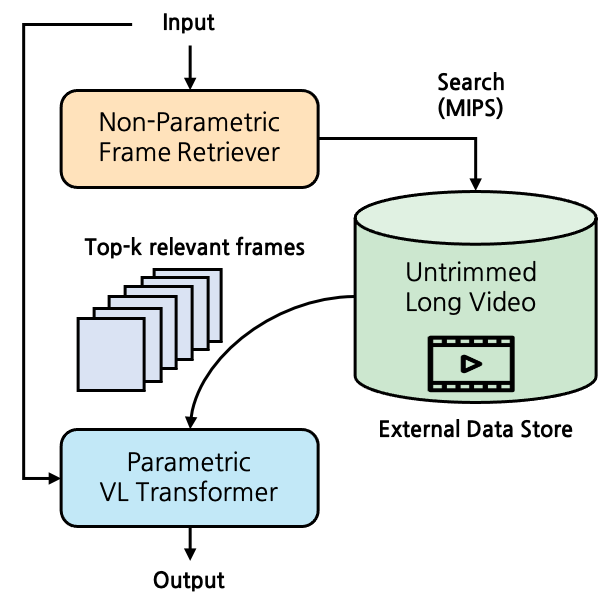}
\caption{Overview of semi-parametric video-grounded text generation. Treating an untrimmed long video as an external data store, it first retrieves top-$k$ relevant frames from the data store with a given input query. Then, a vision-language (VL) transformer encodes each frame and the input independently and generates textual output by performing late fusion over top-$k$ frames.}
\label{fig:motivation}
\vspace{-2mm}
\end{figure}

A line of research has shown the effectiveness of sparse video representation for video-language modeling~\cite{lei2021less}. The sparse video representation approximates a video with sparsely sampled frames instead of all frames while allowing gradient updates of the visual encoder with computationally feasible implementation. \citet{lei2021less} argue that randomly selected sparse frames work well on various video-language tasks even with very few frames, \eg, 1-20 frames for 3-180 seconds clips. Recent video-language studies outperform the performance of the models trained in the single task by massive video-language pre-training, employing the sparse frame paradigm (\eg, uniform sampling)~\cite{zellers2021merlot, wang2022all, zellers2022merlot, yang2022zero}.

However, these studies overlook the limitations of the sparse video representation based on naive frame sampling. The pre-trained models have been tested on only benchmarks with short video clips that are usually less than a minute. We are curious whether the benefits of the sparse frame sampling are still valid if the length of the source video gets longer. In particular, we hypothesize the model relying on a few sampled frames might fail on long untrimmed videos since the scene changes frequently and the frame length affects the size of the population. It requires more frames to be sampled to retain the performance, resulting in an increase in the computational cost, \ie, an efficiency-accuracy trade-off.

On the other hand, recent semi-parametric NLP models show success on knowledge-intensive tasks having a similar challenge regarding large search space of external knowledge~\cite{lewis2020retrieval, izacard2021leveraging}. The semi-parametric models often consist of a non-parametric retriever and a parametric generator. The non-parametric retrieval drastically reduces the search space of large knowledge sources (millions of text such as Wikipedia) to a manageable size, \eg, less than 100, allowing the parametric model to ground relevant knowledge for a given query effectively. Also, it provides controllability of the external knowledge and explainability over model decisions with their provenance. Motivated by their success, we explore \textit{semi-parametric video-grounded text generation} as depicted in Figure~\ref{fig:motivation}, another way for the scalable sparse video representation toward long videos with minutes and even hours.

In this paper, we propose the \textbf{Se}mi-parametric \textbf{Vi}deo-grounded \textbf{T}ext generation model (\frameworkname{}) to take benefits from both efficiency of the sparse frame paradigm and scalability over long-form videos. \frameworkname{} consists of a non-parametric frame retriever and a parametric video-grounded text generator. In particular, we treat a video as an external data store and perform cross-modal retrieval to get top-$k$ query-relevant frames from the data store with a given query. The video-grounded text generator independently encodes each frame with the query. Then, late fusion methods are followed to produce the final output by aggregating the separately encoded query-aware frames, \eg, marginalization in the final decoding step or cross-attention in the decoder layer~\cite{lewis2020retrieval, izacard2021leveraging}.

In our experiments, \frameworkname{} achieves competitive or even better performances on five Video QA~\cite{xu2017video, yu2019activitynet, yang2021just, xiao2021next} and two video captioning~\cite{chen2011collecting, xu2016msr} tasks compared to previous baseline models, which are massively pre-trained on video-text pairs, without any video-language pre-training. Our analysis demonstrates that \frameworkname{} has a significant advantage on the longer videos and questions requiring causal video understanding. Especially, our \frameworkname{} achieves new state-of-the-art performances on three Video QA benchmarks, iVQA, Next-QA, and ActivityQA, which have relatively long source videos, by improving 4.8-6.9\% point of accuracy, and one video captioning, MSRVTT-Caption by improving 3.6\% point of CIDEr.

Our contributions are three folds:
\begin{itemize}

\item To our best knowledge, we propose the semi-parametric architecture in the video-language domain, \frameworkname{}, by treating a video as an external data store for the first time.

\item We demonstrate that \frameworkname{} based on retrieval-augmented generation shows strong performance in long videos and causal video understanding compared to its baseline relying on frame sampling.

\item \frameworkname{} achieves the new state of the art on three Video QA with longer videos, iVQA, Next-QA, Activitynet-QA, and one video captioning, MSRVTT-Caption without any video-language pre-training.
\end{itemize}

%% file: tabs/02_background.tex
\section{Related Work}

\subsection{Video-Language Models}
Previous video-language models~\cite{sun2019videobert, li2020hero, yang2021just} often rely on offline feature extraction leveraging pre-trained 2D/3D vision encoders such as ResNet~\cite{he2016deep}, S3D~\cite{xie2018rethinking} and SlowFast~\cite{feichtenhofer2019slowfast} to efficiently represent video frames, while adopting pre-trained language models like BERT or RoBERTa~\cite{devlin2019bert, liu2019roberta} for the textual representations of subtitles or captions. Recently, some studies adopt end-to-end trainable video-specific transformer~\cite{liu2022video,arnab2021vivit} for video captioning tasks~\cite{lin2022swinbert, seo2022end}.

Contrary to the models relying on the feature extraction for densely sampled frames, \citet{lei2021less} propose ClipBERT representing a video with sparsely sampled frames. It allows end-to-end training of the pre-trained vision and text encoders, leading to comprehensive performances on video-language downstream tasks, \ie, text-to-video retrieval and Video QA.
Recent video-language studies use a few uniformly sampled frames per video to pre-training video-language models~\cite{zellers2021merlot, wang2022all, zellers2022merlot, yang2022zero}. They leverage millions of video-text pairs utilizing automatically generated subtitles via automatic speech recognition API for their pre-training procedure~\cite{miech2019howto100m, bain2021frozen}. The massive pre-training on the large video-text pairs boosts the performance of downstream video-language tasks, achieving state-of-the-art. Our approach shares the sparse frame strategy, but is more scalable toward long videos. Also, we focus on fine-tuning our semi-parametric model rather than video-language pre-training.


\begin{figure*}[t!] 
\centering
\includegraphics[width=0.95\textwidth]{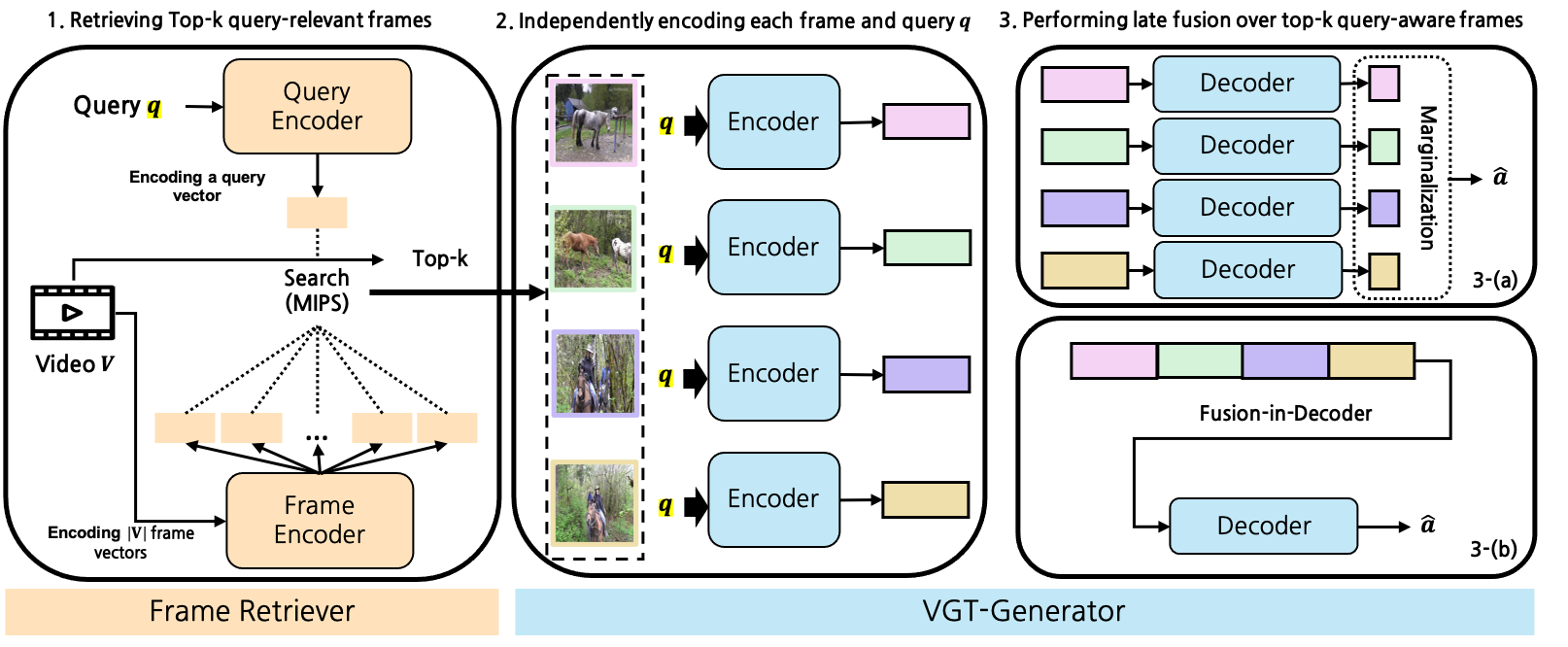}
\caption{Detailed illustration for mechanism of \frameworkname{}. 1) It first retrieves top-$k$ query-relevant frames from a video via maximum inner product search (MIPS) between a query vector and (pre-computed) $|V|$ frame vectors, where $k \ll |V|$. 2) Each frame is encoded with the query $q$ independently by the encoder of VGT-Generator. It produces $k$ query-aware representations. 3) We explore two late fusion methods, 3-(a) Marginalization~\cite{lewis2020retrieval} and 3-(b) Fusion-in-Decoder~\cite{izacard2021leveraging} to produce the final output $\hat{a}$ by aggregating the $k$ query-aware frames in the decoder.}
\label{fig:overview}
\end{figure*}

\subsection{Semi-Parametric Language Models}

Semi-parametric language models show impressive success on many knowledge-intensive NLP tasks such as open-domain question answering and fact varification~\cite{guu2020retrieval, lewis2020retrieval, izacard2021leveraging, izacard2022few}, or language modeling \cite{khandelwal2019generalization, borgeaud2021improving}. The semi-parametric model often consists of a non-parametric module, \ie, a retriever, and a parametric generator. This approach assumes large external knowledge such as Wikipedia or other large text corpora. The non-parametric retriever returns top-$k$ relevant knowledge for a given input from the large data store. The retrieval is often based on a maximum inner product search (MIPS) within pre-computed vectors of the data store~\cite{johnson2019billion}. Then, the parametric generator effectively aggregates the knowledge with the given input. The non-parametric module has several useful properties like controllability, explainability, and debuggability by providing the origin of model decisions. Meanwhile, the parametric model provides better empirical performances than the non-parametric model. Combining the best of two worlds, a semi-parametric architecture is limitedly adopted for protein structure prediction~\cite{jumper2021highly}, image generation~\cite{blattmann2022semi}, and image-text QA~\cite{chen2022murag, lin2022retrieval}. Inspired by these studies, we adapt the retrieval-augmented generation framework to the video-language domain for the first time.

\subsection{Informative Frame Selection}

Focusing on the fact that a video contains a lot of redundant and worthless frames, many works tried to select informative frames from the video~\cite{chen2018less, yu2019activitynet}. \citet{chen2018less} introduce the informative frame selection with the reinforcement learning algorithm for video captioning task. \citet{yu2019activitynet} show informative frame selection leveraging off-the-shelf action proposal network is effective for Video QA with long untrimmed videos. Dense video captioning~\cite{krishna2017dense, zhou2018towards} also contains the frame proposal procedure, but many works evaluate their models with ground-truth proposals~\cite{seo2022end}. On the other hand, \citet{lei2018tvqa} introduce TVQA requiring temporal localization for answering questions in the TV show domain. However, it often needs relevant segment selection with its start and end positions requiring more consideration over corresponding subtitles, \ie, multi-channel tasks. Moreover, the questions containing ``before'' and ``after'' are difficult to find based on the given query because the segment that can infer the answer and the segment that corresponds to the query are often different. We hope our work would be extended with better frame segment selection in future works.

%

%% file: tabs/03_method.tex
\section{Method}

In this section, we introduce \frameworkname{}, a \textbf{Se}mi-parametric \textbf{Vi}deo-grounded \textbf{T}ext generator. As illustrated in Figure~\ref{fig:overview}, it includes informative frame selection leveraging cross-modal retrieval and effective late fusion methods such as marginalization and fusion-in-decoder for video-grounded text generation.
We start by defining task formulation and then explain each method and training details.

\subsection{Overview: Video-Grounded Text Generation}

Let $V = \{f_1, f_2, ..., f_{|V|}\}$ be a video clip consisting of $|V|$ number of sequential frames and $q$ be a textual query. Video-grounded text generation is a task that produces the textual output $\hat{a}$ conditioned on $V$ and $q$, \ie, $p(a \mid V, q)$. Video captioning and video question answering (Video QA) are popular examples of video-grounded text generation. Basically, we inherit the sparse frame paradigm~\cite{lei2021less} representing a video with sparsely selected frames $V_k \subset V$ to approximate a video $V$ where $|V_k| = k$ and $k \ll |V|$. However, in contrast to previous approaches that uniformly select $V_k$, we propose a semi-parametric model which dynamically retrieves the $k$ query-relevant frames using a non-parametric retriever and aggregates them with a parametric generator.

\subsection{Frame Retriever}

Conventionally, many studies perform random uniform sampling to choose $V_k$~\cite{lei2021less,zellers2021merlot, wang2022all,yang2022zero}. In other words, the random sampling selects $k$ frames from $V$ regardless of $q$.


Contrary to the prior studies, we select the relevant frames conditioned on $q$ by introducing a frame retriever. The frame retriever $\eta$ takes $V$ and $q$, and returns a subset $V_k \subset V$, modeled as: $p_\eta(V_k \mid V, q)$. In particular, the frame retriever consists of two separated query and frame transformer encoders, $E_\text{Q}$ and $E_\text{F}$, respectively. Each encoder takes $q$ and $f$ to represent embedding vectors, respectively, where $f_i$ is the $i$-th frame in $V$. Then, the cosine similarity between the two vectors is used for the relevance score between $q$ and $f_i$ as follows:
\begin{equation}
    \text{sim}(q, f_i) = \frac{E_\text{Q}(q)^T E_\text{F}(f_i)}{\|E_\text{Q}(q)\|_2 \|E_\text{F}(f_i)\|_2}
    \label{eq:cosine_sim}
\end{equation}
where $\|\cdot||_2$ denotes the $l$-2 normalization.
Frame retriever returns the top-$k$ frames based on the relevance scores among $q$ and all frames in $V$ as follows:
\begin{equation}
    V_k \leftarrow \argsort_{f_i \in V}(\text{sim}(q, f_i))[:k].
    \label{eq:argsort}
\end{equation}

Also, we compute the relative importance of each selected frame $f_j \in V_k$ by performing softmax over the cosine similarities (Equation~\ref{eq:cosine_sim}) where $\tau$ is a temperature hyper-parameter. The frame score is computed as follows:

\begin{equation}
    p_{\eta}(f_j \mid q) = \frac{e^{\text{sim}(f_j, q)/\tau}}{\sum_k e^{\text{sim}(f_k, q)/\tau}}
    \label{eq:frame_score}
\end{equation}

\subsection{Video-Grounded Text Generator}

A video-grounded text (VGT) generator $\theta$ takes $V_k$ and $q$, and it outputs $a$. For $\theta$, we leverage the transformer encoder-decoder architecture taking both image and text together to generate textual output~\cite{vaswani2017attention, wang2021simvlm, wang2022unifying}. Specifically, it first embeds each frame $f_j \in V_k$ and a text query $q$ with convolution blocks such as ResNet~\cite{he2016deep} and embedding matrix lookup corresponding to subwords from byte-pair encoding~\cite{sennrich2015neural}, respectively. Then, the frame patches and subword tokens vectors are combined and fed into the multi-modal transformer encoder to produce $k$ query-aware frame representations. Beyond the single frame and query interaction, we investigate two effective late fusion methods, Marginalization~\cite{lewis2020retrieval} and Fusion-in-Decoder~\cite{izacard2021leveraging}, to aggregate the independently encoded $k$ query-aware frames for generating target text $a$ in the decoder.

\paragraph{Marginalization (MAR)}

It integrates the $k$ query-aware frames by marginalization~\cite{lewis2020retrieval}. First, the decoder also produces independent $k$ predictions. Then, it aggregates the $k$ predictions by marginalizing out weighting by the frame score $p_{\eta}(f_j \mid q)$ resulting in the output $a = \{w_1, w_2, ..., w_N\}$, where the $w$ is a subword token of $a$.
\begin{equation}
    p(a \mid V, q) = \prod_i^N \sum_{f_j \in V_k} p_{\eta}(f_j \mid q) p_{\theta}(w_i \mid q, f_j, w_{1:i-1})
    \label{eq:rag}
\end{equation}

The marginalization procedure allows joint optimization of the cross-modal retriever and generator. In other words, it enables gradient updates of encoders in a cross-modal retriever to select query-relevant frames with Equation~\ref{eq:rag}, while not requiring explicit supervision for ground-truth query-relevant frame pairs.

\paragraph{Fusion-in-Decoder (FiD)}

Fusion-in-Decoder relies purely on cross-attention between the hidden states of the encoder and decoder for the fusion~\cite{izacard2021leveraging}. Like the marginalization, it encodes $q$ with each $f_j \in V_k$ independently. However, it aggregates the encoder outputs jointly in the decoder with cross-attention as illustrated in Figure~\ref{fig:overview}. Specifically, the encoder produces the hidden states $H \in \mathbb{R}^{k \times L \times d}$, where the $L$ is the length of the combined frame and query outputs, and the $d$ is a hidden dimension. The $k$ hidden outputs are concatenated together as $H \in \mathbb{R}^{k \cdot L \times d}$, as a single sequence before being fed into the decoder. Finally, the decoder can consider the $k$ query-aware frames at the same time for target text generation.
\begin{equation}
    p(a \mid V, q) = \prod_i^N p_{\theta}(w_i \mid q, V_k, w_{1:i-1})
    \label{eq:fid}
\end{equation}

\subsection{Training}
\label{sec:sevit_training}

VGT-generator is trained by minimizing the negative log-likelihood of $p(a \mid V, q)$ with either Equation~\ref{eq:rag} or \ref{eq:fid}. For efficient implementation, we pre-compute the frame vectors in advance for all training videos of the target dataset using the frame encoder of the frame retriever. Then, an efficient search algorithm, \ie, Maximum Inner Product Search (MIPS), becomes convenient especially when the length of the source video gets longer.
We further describe some training techniques considering the frame retriever.

\paragraph{Query-side Fine-Tuning}

With the objective of Marginalization (Equation~\ref{eq:rag}), we can jointly optimize the frame retriever and VGT-generator. However, re-computation of frame vectors is required for all videos when we update the frame encoder of the frame retriever. \citet{lewis2020retrieval, izacard2022few} report that the updating context encoder does not show significant advantages for knowledge-intensive NLP tasks despite the heavy computation. Thus, we keep the frame encoder $E_\text{F}$ fixed during the training while only updating the query encoder $E_\text{Q}$ for efficiency~\cite{lewis2020retrieval, izacard2022few}.

\paragraph{Retriever Warm-up for FiD}

On the other hand, joint training of the frame retriever with the objective of FiD is not straightforward. Even though \citet{izacard2022few} propose various methods for joint training of retriever and generator in FiD fusion scheme, it does not work well for ours in our preliminary experiments. Thus, we initialize the frame retriever with a fine-tuned retriever by Marginalization. Then, we fix the frame retriever during training VGT-generator with the FiD manner. It is a similar approach to FiD-RAG in \citet{shuster2021retrieval}.

\paragraph{Top-k Annealing}

We promote diverse top-$k$ frame selection with the fixed retriever in FiD training. Basically, we choose a frame from $V$ in order of high relevance score (Equation~\ref{eq:argsort}). However, VGT-generator might show a lower generalization ability if trained with the same $k$ frames from the fixed retriever for every training instance. Thus, we set a window size $u \in \mathbb{R}^1$ and prevent subset $\{f_{i-u}, f_{i-u+1}, ..., f_i, ..., f_{i+u-1}, f_{i+u}\}$ from being selected once $f_i$ is selected as one of top-$k$ frames. We gradually decrease the $u \rightarrow 0$ at every training epoch, resulting in diverse top-$k$ frames.

%% file: tabs/04_experimental_setup.tex
\section{Experiments}

In this section, we demonstrate the effectiveness of \frameworkname{} compared to its baselines on eight video-language datasets. We denote our models as \frameworkname{}$_\text{MAR}$ and \frameworkname{}$_\text{FiD}$ according to their training objectives, Marginalization and Fusion-in-Decoder, respectively. 

\input{tables/fusion_methods.tex}

\subsection{Main Baseline: \frameworkname{} with Frame Sampling}
\label{subsec:baseline}

Although there are several baseline models, we would like to strictly compare the effect of employing the frame retrieval while controlling other factors such as model size, pre-training steps, and other fusion methods. To this end, we introduce a strong baseline utilizing uniform frame sampling instead of performing frame retrieval to choose $k$ frames. We refer to the baseline as \frameworkname{}-with-frame-sampling, denoting it simply \frameworkname{}$^{\otimes}$. When we train the \frameworkname{}$^{\otimes}$ with the Marginalization, we use a uniform prior $1/k$ for the frame score per frame instead of using Equation~\ref{eq:frame_score} for the late fusion (Equation~\ref{eq:rag}). We describe details of other baselines in our experiments in Appendix~\ref{appendix:baseline}.

\subsection{Dataset}

We evaluate our models on six Video QA datasets, TGIF-QA~\cite{jang2017tgif}, MSVD-QA~\cite{xu2017video}, MSRVTT-QA~\cite{xu2017video}, iVQA~\cite{yang2021just}, Next-QA~\cite{xiao2021next}, and Activitynet-QA~\cite{yu2019activitynet}, and two video captioning datasets, MSVD-Caption~\cite{chen2011collecting} and MSRVTT-Caption~\cite{xu2016msr}. We mainly report top-1 accuracy for the Video QA and CIDEr~\cite{vedantam2015cider} for the video captioning. More details including statistics are in Appendix~\ref{appendix:dataset}.

\subsection{Implementation Details}

We use pre-trained CLIP-base/16~\cite{radford2021learning} for our frame retriever and pre-trained OFA-Base~\cite{wang2022unifying} for the VGT-generator. CLIP is a pre-trained bi-encoder on large image-text pairs. OFA is a pre-trained vision-language transformer on multi-tasks consisting of image-text, text-only, and image-only tasks, by unifying the input and output protocol. As described in Section~\ref{sec:sevit_training}, we pre-compute frame vectors of all videos in the target dataset in advance to perform an efficient search, MIPS. The temperature $\tau$ is empirically set to 1. First, we train \frameworkname{}$_\text{MAR}$ in the Marginalization, \ie, joint optimization of the retriever and generator on the target dataset. Then, the fine-tuned retriever is reused for training \frameworkname{}$_\text{FiD}$ on the same dataset as described in Section~\ref{sec:sevit_training}. Also, we set $k$ to 5 for training and 10 at test time for all datasets except for TGIF-QAs where we set $k$ to 3 and 6 for the training and test.
For multiple-choice QA, TGIF-Action, TGIF-Transition, and Next-QA, we concatenate answer options and query together introducing a separation token. For video captioning tasks, we use a null query, ``\textit{What does the image describe?}'', used for image captioning tasks by \citet{wang2022unifying}. All models are trained with \{1e-5, 3e-5\} learning rate and \{16, 32\} batch size for 5 epochs on 1-2 NVIDIA A100 GPUs. We use mainly PyTorch and Huggingface's Transformers library for our implementation~\cite{paszke2019pytorch, wolf2020transformers}. Please see Appendix~\ref{appendix:implementation} for more details.

%% file: tables/fusion_methods.tex
\begin{table}[t!]
    \centering
    \small
    \caption{Comparison between two fusion methods, Marginalization (MAR) and Fusion-in-Decoder (FiD) on Video QA and captioning tasks based on our baseline with uniform frame sampling, \frameworkname{}$^{\otimes}$ explained in Section~\ref{subsec:baseline}, to identify their differences apart from frame retrieval. We report top-1 accuracy for Video QA and CIDEr for video captioning.}
    \vskip 0.15in
    \begin{tabular}{lccc}
        \toprule
        Dataset & \makecell{\# Frame \\ train/test} & MAR & FiD \\
        \midrule
        \textit{Video QA} & & & \\
        TGIF-Action & 3/6 & \textbf{94.9} & 94.8 \\
        TGIF-Transition & 3/6 & \textbf{98.2} & 98.0 \\
        TGIF-Frame & 3/6 & 70.6 & \textbf{71.1} \\
        MSVD-QA & 5/10 & \textbf{49.3} & \textbf{49.3} \\
        MSRVTT-QA & 5/10 & 41.9 & \textbf{42.3} \\
        iVQA & 5/10 & 35.3 & \textbf{36.4} \\
        Next-QA & 5/10 & 54.4 & \textbf{54.6} \\
        Activitynet-QA & 5/10 & 46.3 & \textbf{47.1} \\
        \midrule
        \textit{Video Captioning} & & & \\
        MSVD-Caption & 5/10 & 127.4 & \textbf{134.9} \\
        MSRVTT-Caption & 5/10 & 58.6 & \textbf{61.8} \\
        \bottomrule
    \end{tabular}
    \label{table:comparing_fusion_method}
    \vspace{-2mm}
\end{table}

%% file: tabs/05_experimental_result.tex
\input{tables/frame_retrieval.tex}
\input{tables/long_subsplit.tex}
\input{tables/nextqa_val.tex}

\subsection{Comparison between Late Fusion Methods}


Before we discuss the benefits of frame retrieval, we compare our two late fusion methods based on our baseline model, \frameworkname{}$^{\otimes}$. Table~\ref{table:comparing_fusion_method} shows the results on ten downstream datasets. Both methods show comparable performance to each other, but FiD performs slightly better than MAR in most Video QA datasets, especially in TGIF-Frame, iVQA, and Activitynet-QA. Those datasets contain descriptive QA pairs. We find that the late fusion methods perform surprisingly well on the datasets requiring temporal reasoning, \eg, TGIF-Action, TGIF-Transition, and Next-QA, even though they do not consider the temporal order among frames explicitly. Also, FiD shows better performances than MAR in the two video captioning tasks indicating FiD is good at generating longer text.

\input{tables/activitynet_breakdown.tex}

\subsection{Benefits from Frame Retrieval}

Table~\ref{table:comparing_selection_method} shows the effect of frame retrieval on the Video QA datasets~\footnote{We exclude TGIF-QAs since their video length (3s) makes our frame retrieval meaningless.}. The frame retrieval consistently improves the performances of all datasets except for MSRVTT-QA, regardless of video length. Notably, it improves the performances of longer Video QA datasets, iVQA, Next-QA, and Activitynet-QA. We find that gains are slightly larger in MAR fusion improving the 1.0 and 0.9 accuracies of iVQA and Activitynet-QA, respectively. We presume the joint training of frame retrieval boosts the gain. In Table~\ref{table:captioning} of Appendix~\ref{appendix:experiments}, we find that frame retrieval consistently improves performances of video captioning as well, even though the null query is used for the retrieval. We think the null query effectively filters out uninformative frames resulting in performance gains.

\paragraph{Results on Long Video Subset}

In Table~\ref{table:long-subsplit}, we further break down the evaluation results according to source video length to identify the benefits of frame retrieval in longer videos. Specifically, we divide the original test set of Next-QA and Activitynet-QA into long and short sub-splits according to whether the source video length is longer than 60 seconds. We can find that most improvements by frame retrieval are from the long videos in both datasets. Especially, \frameworkname{}$_\text{FiD}$ improves 3.2\% point accuracy of long video subset in Next-QA by employing frame retrieval.

\input{tables/ablation_study.tex}

\paragraph{Results by Question Types}

Table~\ref{table:nextqa} shows that the performance gains by the frame retrieval in Next-QA are related to causal and temporal QA types for both fusion methods. In contrast, the performance of the descriptive type is slightly degraded by the frame retrieval. It is notable that the improvements are significant in \frameworkname{}$_\text{MAR}$ for both causal and temporal types while the improvement is limited to causal type in \frameworkname{}$_\text{FiD}$. It reminds us of the importance of joint retriever training, again. However, \frameworkname{}$_\text{FiD}$ consistently performs better in all three types compared to \frameworkname{}$_\text{MAR}$. Moreover, it outperforms previous best-performing models utilizing sophisticated graph representation, HGA~\cite{jiang2020reasoning}, HGQA~\cite{xiao2022video}, and VGT~\cite{xiao2022videotransformer}, especially in causal and descriptive types. More results by question types are in Table~\ref{table:activitynet_question} and~\ref{table:nextqa_finegrained} of Appendix~\ref{appendix:experiments}.

\input{tables/videoqa_sota_comparison.tex}

\paragraph{Analysis on Untrimmed Long Videos}

Finally, we further analyze the advantages of frame retrieval on the longest video benchmark, Activitynet-QA. Figure~\ref{fig:activitynet} illustrates the advantages in two folds. First, we divide the test set into subsets according to more fine-grained video lengths as shown in Figure~\ref{fig:activitynet} (a). As hypothesized, there is a clear tendency for significant performance gaps according to the usage of frame retrieval to become more pronounced in longer videos. Specifically, \frameworkname{}$^{\otimes}_\text{MAR}$ and \frameworkname{}$^{\otimes}_\text{FiD}$ both drop their performances significantly with videos longer than 180 seconds. However, both \frameworkname{}$_\text{MAR}$ and \frameworkname{}$_\text{FiD}$ successfully retain their performances with the longer videos. Second, we investigate the sample efficiency in terms of the number of frames at the inference time. We believe that if the frame retriever selects informative frames well, the model works better with fewer frames than non-informative frames from the uniform frame sampler. Figure~\ref{fig:activitynet} (b) shows such a tendency as we have hypothesized. Even though the performances decrease gradually with fewer frames, the performance gaps between \frameworkname{} and \frameworkname{}$^{\otimes}$ also becomes significant. It implies the frames obtained by frame retriever are more informative than the frames by random uniform sampling. We also find the strength of \frameworkname{} in long videos by a qualitative analysis as shown in Figure~\ref{fig:qualitative_example} of Appendix~\ref{appendix:qualitative}.

\subsection{Ablation Study}

Table~\ref{table:ablation_study} shows ablation results on two long Video QA datasets, Next-QA and Activitynet-QA. Once again, we can observe the importance of frame retrieval from this study. If we fix the frame retriever during training instead of performing query-side (QS) fine-tuning, the final performance of \frameworkname{}$_\text{MAR}$ drops 0.6 and 1.2 of accuracy in Next-QA and ActivitynetQA, respectively. Similarly, using the warmed-up frame retriever by \frameworkname{}$_\text{MAR}$ boosts the final performance of \frameworkname{}$_\text{FiD}$  in both datasets. We find diverse retrieval by top-k annealing also contributes to the final performance. Furthermore, we also find the larger backbone model for the VGT-generator significantly improves performance.

\subsection{Comparison with State-of-the-arts}

We also compare ours with previous state-of-the-art models as shown in Table~\ref{table:main_result}. Notably, our models show competitive performances on the short video-based Video QA datasets compared to baseline models pre-trained on large video-text pairs, JustAsk~\cite{yang2021just}, MERLOT~\cite{zellers2021merlot}, All-in-One~\cite{wang2022all}, FrozenBiLM~\cite{yang2022zero} and LAVENDER~\cite{li2022lavender}, even without any video-text pre-training. Also, our model outperforms other baselines utilizing graph representation, HQGA~\cite{xiao2022video}, IGV~\cite{li2022invariant}, and VGT~\cite{xiao2022videotransformer}, and baselines pre-trained on image-text pairs, ClipBERT~\cite{lei2021less} and SINGULARITY~\cite{lei2022revealing}. Moreover, our models outperform in relatively longer Video QA datasets, Next-QA and Activitynet-QA. Especially, our FiD-based model achieves new state-of-the-art performances on iVQA, Next-QA, and Activitynet-QA, when using a large-sized backbone, \ie, OFA-Large, for our VGT-generator. Moreover, in Table~\ref{table:captioning} of Appendix~\ref{appendix:experiments}, our model shows competitive performances on the video captioning dataset compared to end-to-end video transformer-based baselines, SwinBERT~\cite{lin2022swinbert}, MV-GPT~\cite{seo2022end}, and LAVENDER~\cite{ li2022lavender}. Notably, \frameworkname{}$_\text{FiD}$ based on OFA-Large achieves a new state-of-the-art performance in terms of CIDEr~\cite{vedantam2015cider} on the MSRVTT-Caption dataset even without video-text pre-training.

%% file: tables/frame_retrieval.tex
\begin{table}[t!]
    \centering
    \small
    \caption{We compare our \frameworkname{} leveraging frame retriever with its counterpart baseline, \frameworkname{}$^{\otimes}$ relying on uniform frame sampling instead of frame retrieval, on Video QA datasets.}
    \vskip 0.15in
    \begin{tabular}{lccc}
        \toprule
        \makecell{Dataset \\ \small{(Avg. video length)}} & \makecell{Frame \\ Retrieval} & MAR & FiD \\
        \midrule
        \multirow{2}{*}{MSVD-QA \small{(10s)}} & \xmark & 49.3 & 49.3 \\
        & \cmark & \textbf{49.5} & \textbf{49.7} \\
        \midrule
        \multirow{2}{*}{MSRVTT-QA \small{(15s)}} & \xmark & \textbf{41.9} & \textbf{42.3} \\
        & \cmark & 41.7 & 42.1 \\
        \midrule
        \multirow{2}{*}{iVQA \small{(18s)}} & \xmark & 35.3 & 36.4 \\
        & \cmark & \textbf{36.4} & \textbf{36.9} \\
        \midrule
        \multirow{2}{*}{Next-QA \small{(44s)}} & \xmark & 54.4 & 54.6 \\
        & \cmark & \textbf{54.8} & \textbf{55.2} \\
        \midrule
        \multirow{2}{*}{Activitynet-QA \small{(180s)}} & \xmark & 46.3 & 47.1 \\
        & \cmark & \textbf{47.2} & \textbf{47.6} \\
        \bottomrule
    \end{tabular}
    
    \label{table:comparing_selection_method}
\end{table}

%% file: tables/long_subsplit.tex
\begin{table}[t!]
    \centering
    \small
    \caption{Evaluation breakdown on Next-QA and Activitynet-QA by their source video length. We split the test set of Next-QA and Activitynet-QA into long and short subsets according to whether the source video length is longer than 60 seconds.}
    \vskip 0.15in
    \begin{tabular}{lcccccc}
        \toprule
        \multirow{2}[3]{*}{Model} & \multicolumn{3}{c}{Next-QA} & \multicolumn{3}{c}{Activitynet-QA} \\
        \cmidrule(r){2-4}
        \cmidrule(r){5-7}
        & Short & Long & All & Short & Long & All \\
        \midrule
        \frameworkname{}$_\text{MAR}^{\otimes}$ & 54.6 & 53.4 & 54.4 & 47.3 & 45.9 & 46.3 \\
        \frameworkname{}$_\text{MAR}$ & \textbf{55.1} & \textbf{53.9} & \textbf{54.8} & \textbf{47.4} & \textbf{47.1} & \textbf{47.2} \\
        \midrule
        \frameworkname{}$_\text{FiD}^{\otimes}$ & \textbf{54.9} & 53.4 & 54.6 & \textbf{47.9} & 46.6 & 47.1 \\
        \frameworkname{}$_\text{FiD}$ & 54.8 & \textbf{56.6} & \textbf{55.2} & 47.5 & \textbf{47.6} & \textbf{47.6} \\
        \bottomrule
    \end{tabular}
    
    \label{table:long-subsplit}
\end{table}

%% file: tables/nextqa_val.tex
\begin{table}[t!]
    \centering
    \small
    \caption{Evalution results on Next-QA validation set. In addition to the original validation set, we include a hard subset requiring video-level understanding identified by ATP~\cite{buch2022revisiting}.}
    \vskip 0.15in
    \vskip 0.15in
   \begin{adjustbox}{width=0.48\textwidth} 
    \begin{tabular}{lcccc}
        \toprule
        Model & \makecell{Causal \\ \scriptsize{Original / Hard}} & \makecell{Temporal  \\ \scriptsize{Original / Hard}} & Descriptive & All \\
        \midrule
        HGA & 46.3 / 43.3 & 50.7 / 45.3 & 59.3 & 49.7 \\
        HGQA & 48.5 / $\;\;\,$-$\;\;\,$ & 51.2 / $\;\;\,$-$\;\;\,$ & 61.7 & 51.4 \\
        APT & 48.3 / 19.6 & 46.7 / 22.6 & 58.9 & 49.2 \\
        Temp[APT] & 48.6 / 38.4 & 49.3 / 36.5 & 65.0 & 51.5 \\
        \quad + APT & 53.1 / $\;\;\,$-$\;\;\,$ & 50.2 / $\;\;\,$-$\;\;\,$ & 66.8 & 54.3 \\
        VGT & 52.3 / $\;\;\,$-$\;\;\,$ & \textbf{55.1} / $\;\;\,$-$\;\;\,$ & 64.1 & 55.0 \\
        \midrule
        \frameworkname{}$_\text{MAR}^{\otimes}$ & 52.3 / 41.9 & 52.3 / 44.7 & 71.2 & 55.2 \\
        \frameworkname{}$_\text{MAR}$ & 53.5 / 43.2 & 54.0 / 46.3 & 69.2 & 56.1 \\
        \frameworkname{}$_\text{FiD}^{\otimes}$ & 53.0 / 42.7 & 54.1 / 46.4 & \textbf{71.9} & 56.3 \\
        \frameworkname{}$_\text{FiD}$ & \textbf{54.0} / \textbf{43.3} & 54.1 / \textbf{46.5} & 71.3 & \textbf{56.7} \\
        \bottomrule
    \end{tabular}
    \end{adjustbox}
    \label{table:nextqa}
\end{table}

%% file: tables/activitynet_breakdown.tex
\begin{figure*}[t!] 
\centering
\includegraphics[width=0.95\textwidth]{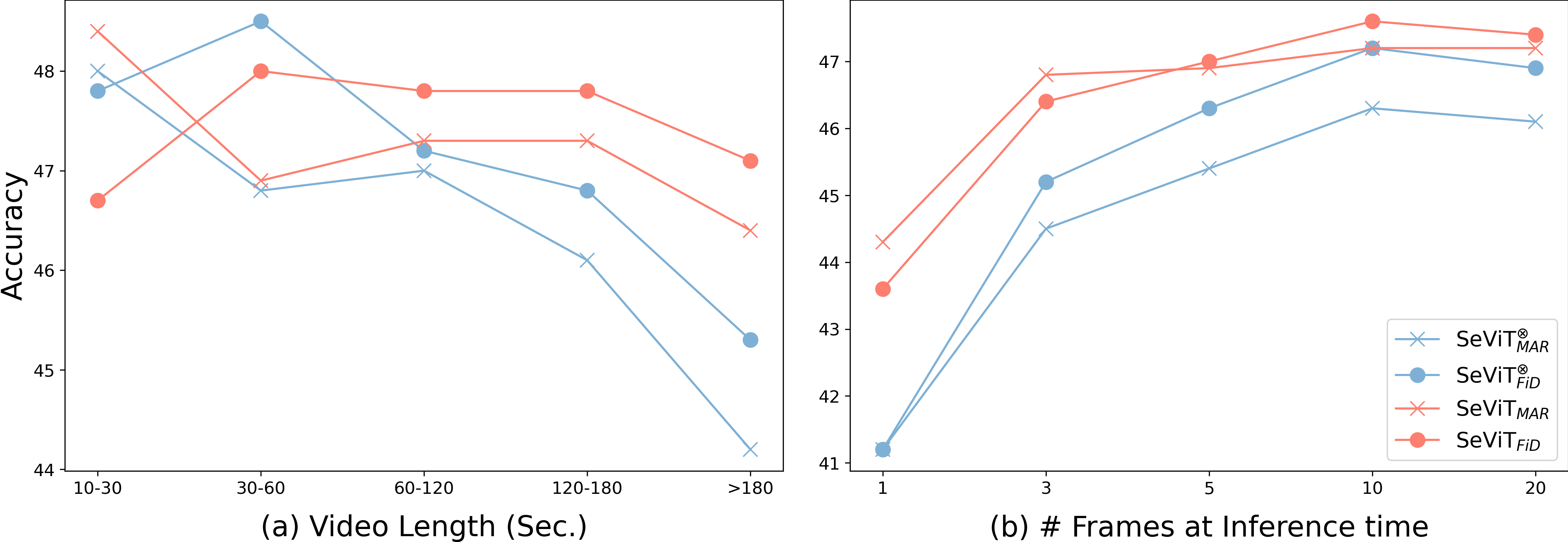}
\caption{Breakdown results on Activitynet-QA~\cite{yu2019activitynet} by (a) source video length and (b) the number of frames at the test time.}
\label{fig:activitynet}
\end{figure*}

%% file: tables/ablation_study.tex
\begin{table}[t!]
    \centering
    \small
    \caption{Ablation study on two long Video QA datasets, Next-QA and Activitynet-QA.}
    \vskip 0.15in
    \begin{tabular}{lcc}
        \toprule
        Model & Next-QA & Activitynet-QA \\
        \midrule
        \frameworkname{}$_\text{MAR}$ & \textbf{54.8} & \textbf{47.2} \\
        \quad w/o. QS fine-tuning & 54.2 & 46.0 \\
        \midrule
        \frameworkname{}$_\text{FiD}$ & \textbf{55.2} & \textbf{47.6} \\
        \quad w/o. Retriever warm-up & 54.4 & 47.0 \\
        \quad w/o. Top-k annealing & 54.3 & 46.7 \\
        \midrule
        \frameworkname{}$_\text{FiD}$ & &  \\
        \quad w. OFA-Medium \small{(93M)} & 51.1 & 44.8 \\
        \quad w. OFA-Base \small{(182M)} & 55.2 & 47.6 \\
        \quad w. OFA-Large \small{(472M)} & \textbf{60.6} & \textbf{48.9} \\
        
        \bottomrule
    \end{tabular}
    
    \label{table:ablation_study}
\end{table}

%% file: tables/videoqa_sota_comparison.tex
\begin{table*}[t!]
    \centering
    \small
    \caption{Comparison with baselines including state-of-the-art models on five Video QA datasets. $\dagger$ indicates our VGT-generator is initialized with OFA-Large~\cite{wang2022unifying}. $*$ indicates the score is obtained from \citet{xiao2022videotransformer}. \textbf{Bold} indicates the best score and \underline{underline} indicates the second best score.}
    \vskip 0.15in
    \begin{tabular}{lcccccc}
        \toprule
        Model & \makecell{Video-Text \\
        Pre-training} & \makecell{MSVD-QA \\ (10s)} & \makecell{MSRVTT-QA \\ (15s)} & \makecell{iVQA \\ (18s)}& \makecell{Next-QA \\ (44s)} & \makecell{Activitynet-QA \\ (180s)} \\
        \midrule
        JustAsk & \cmark & 47.5 & 41.8 & 35.4 & 50.8$^*$ & 39.0 \\
        MERLOT & \cmark & - & 43.1 & - & - & 41.4 \\
        All-in-One & \cmark & 48.3 & \underline{46.8} & - & - & - \\
        FrozenBiLM & \cmark & \underline{54.4} & \textbf{47.0} & \underline{39.7} & - & 43.2 \\
        LAVENDER& \cmark & \textbf{56.6} & 45.0 & - & - & - \\
        \midrule
        ClipBERT & \xmark & - & 37.4 & - & - & - \\
        SINGULARITY & \xmark & - & 43.9 & - & - & 44.1 \\
        IGV & \xmark & 40.8 & 38.3 & - & 51.3 & - \\
        HQGA & \xmark &  41.2 & 38.6 & - & 51.8 & - \\
        VGT & \xmark & - & 39.7 & - & 53.7 & - \\
        \midrule
        \textit{Ours} & & & & & & \\
        \frameworkname{}$_\text{MAR}$ & \xmark & 49.5 & 42.3 & 36.4 & 54.8 & 47.2 \\
        \frameworkname{}$_\text{FiD}$ & \xmark & 49.7 & 42.1 & 36.9 & \underline{55.2} & \underline{47.6} \\
        \midrule
        \frameworkname{}$_\text{FiD}^{\dagger}$ & \xmark & 52.6 &  43.8 & \textbf{44.5} & \textbf{60.6} & \textbf{48.9} \\
        \bottomrule
    \end{tabular}
    \label{table:main_result}
\end{table*}

%% file: tabs/06_conclusion.tex
\section{Conclusion}

In this work, we present \frameworkname{} for scalable video representation toward untrimmed long videos. In particular, we regard a video as an external data store and leverage the non-parametric retriever to get relevant frames. Then, a parametric generator focuses on the effective aggregation of the frames. We find \frameworkname{} has significant advantages, especially in longer videos and questions requiring causal video understanding. Furthermore, \frameworkname{} achieves state-of-the-art performances on Video QA and captioning tasks without any video-language pre-training. We believe \frameworkname{} will promote future research into longer video understanding, \eg, minutes or even hours.

%% file: tabs/08_appendix.tex
\clearpage
\appendix

\section{Dataset Details}
\label{appendix:dataset}

TGIF-QAs include three types of spatio-temporal Video QA based on short video clips (GIFs) of 3 seconds on average~\cite{jang2017tgif}. TGIF-Action and TGIF-Transition provide candidate answer options, while TGIF-Frame is an open-ended QA. MSVD-QA and MSRVTT-QA are also widely used open-ended Video QA datasets containing short video clips of 10-15 seconds on average and auto-constructed synthetic questions~\cite{xu2017video}. iVQA~\cite{yang2021just} is an open-ended Video QA dataset with video clips of 18 seconds on average and human annotations of 5 reference answers per question. We follow the weighted accuracy evaluation setup of \citet{yang2021just}. Recently, \citet{xiao2021next} introduce Next-QA to evaluate video understanding focusing on causal and temporal aspects of daily activity contents~\cite{xiao2021next}. It is a multiple-choice QA dataset including relatively longer videos with 44 seconds on average. We also report validation results on hard sub-split of Next-QA requiring video-level understanding identified by \citet{buch2022revisiting}. Activitynet-QA contains human-generated QA pairs based on the longest untrimmed videos with 180 seconds on average~\cite{yu2019activitynet}. MSVD-Caption and MSRVTT-Caption datasets provide 40 and 20 ground-truth captions based on 10-15 seconds video clips, respectively~\cite{chen2011collecting, xu2016msr}. For all datasets, we compress videos with \{0.5, 1\} FPS to increase processing efficiency. The statistics of datasets are shown in Table~\ref{table:dataset}.

\input{tables/dataset.tex}

\section{More Implementation Details}
\label{appendix:implementation}

\input{tables/hyperparameter.tex}

We set the maximum sequence length of the query to 32 for open-ended Video QA and video captioning tasks, and 64 for multiple-choice Video QA datasets. Also, we set the beam size to 6 and no-repeat-ngram-size to 3 for the generation. All hyper-parameter setups are in Table~\ref{table:common_hyper} and Table~\ref{table:specific_hyper}.
For all experiments, we choose the best-performing checkpoint based on the validation score.

\section{Other Baselines}
\label{appendix:baseline}

\paragraph{Video QA}

\citet{lei2021less} propose \textbf{ClipBERT}, an end-to-end video-language model with sparse frame sampling. Contrary to previous approaches relying on pre-extracted dense video features, it leverages sparsely sampled frames as a video representation allowing gradient updates. Also, \citet{lei2021less} conduct image-text pre-training based on BERT~\cite{devlin2019bert} and ResNet-50~\cite{he2016deep} backbones. It shows the benefits of sparse sampling in many video-language downstream tasks. On the other hand, \citet{lei2022revealing} explain single frame bias in video-language tasks by introducing \textbf{SINGULARITY} model. The authors argue that some video-language tasks do not require reasoning over multiple frames. Rather, they claim that it is enough for the tasks to perform the proper fusion method over frames at test time, after image-text pre-training and fine-tuning.

\citet{yang2021just} introduce synthetic Video QA dataset, HowToVQA69M, by generating question-answer pairs conditioned on only transcripts. They show the model pre-trained on the synthetic dataset works well on various Video QA datasets. The resulting model is VQA-T, which we refer to \textbf{JustAsk}. The model utilizes offline features from DistillBERT~\cite{sanh2019distilbert} and S3D~\cite{xie2018rethinking} for text and video representation, respectively. It is optimized by contrastive loss, between the video-question pair and the correct answer, and masked language modeling (MLM) loss. \textbf{MERLOT}~\cite{zellers2021merlot} is another video-language model pre-trained on the YT-Temporal-180M, the curated video-transcript dataset from 6M unlabelled YouTube videos. They employ a vision transformer on the top of ResNet-50~\cite{he2016deep} to represent video with a few sampled frames. It is trained by optimizing contrastive frame-text matching, masked language modeling (MLM), and temporal reordering objectives. They demonstrate the efficacy of the large pre-training on various visual reasoning and Video QA tasks. Similarly, \textbf{All-in-One} adopts MLM and the frame-text matching loss for its pre-training objectives~\cite{wang2022unifying}. Specifically, it unifies the input representation that embeds raw pixels and text jointly. Moreover, it also relies on sparsely sampled frames, \eg, 3 or 9 frames. \textbf{FrozenBiLM} is devised for zero-shot Video QA by \citet{yang2021just}. They conduct parameter-efficient tuning by introducing a few trainable parameters in frozen pre-trained bidirectional encoder~\cite{he2021deberta}. It is further pre-trained on 10M video-text pairs with MLM objective. The model achieves state-of-the-art performances on many (zero-shot) Video QA benchmarks.

On the other hand, \textbf{HGA}~\cite{jiang2020reasoning} performs cross-modal reasoning over heterogeneous graphs among the video frames and question words for reach modality interaction. \citet{xiao2021next} test the model in their benchmark, Next-QA, with BERT~\cite{devlin2019bert} for the text representations. \textbf{HQGA}~\cite{xiao2022video} represents a video as a hierarchical semantics consisting of multiple granularities, \eg, entity, frame, clip, and video, with textual cues. It shows good performance on the Video QA such as Next-QA. Similarly, \textbf{VGT}~\cite{xiao2022videotransformer} also utilizes the video graph representation explicitly capturing objects in the frames, their relationships, and spatio-temporal dynamics. \citet{li2022invariant} introduce \textbf{IGV} with a training framework to prevent Video QA models from exploiting spurious correlations. In particular, they split a video into causal and complement parts with respect to the given question. Then, the Video QA models are trained to exploit causal frames while not grounding on the complement frames which do not contain critical clues.

\paragraph{Video Captioning}

\textbf{ORG-TRL}~\cite{zhang2020object} employs graph convolutional networks to model objects in the video and conducts teacher-recommended learning leveraging pre-trained language model~\cite{devlin2019bert}. \citet{tang2021decembert} introduce \textbf{DECEMBERT} pre-trained by noisy video-transcript pairs along with corresponding dense captions from the off-the-shelf dense captioning model. They pre-train the model with MLM and video-text matching loss. Moreover, they introduce constrained attention loss to mitigate misalignment errors of noisy transcripts.
\citet{lin2022swinbert} propose a fully end-to-end trainable transformer, \textbf{SwinBERT} on the top of VidSwin Transformer~\cite{liu2022video} for video captioning. To handle long video input efficiently, \citet{lin2022swinbert} introduce learnable sparse attention masks that reduce redundant video inputs. \textbf{LAVENDER} also adopts the VidSwin as a backbone and unifies the formulation of pre-training and fine-tuning with MLM~\cite{li2022lavender}. \citet{seo2022end} propose \textbf{MV-GPT} which is another end-to-end video-language model for video captioning. They employ the video-specific transformer backbone, ViViT~\cite{arnab2021vivit} to represent a video. They also introduce a bidirectional pre-training objective predicting either present or future subtitles from masked text input and corresponding frames.

\section{More Experimental Results}
\label{appendix:experiments}

We further report breakdown results by question types on Activitynet-QA test set in Table~\ref{table:activitynet_question} and Next-QA validation set in Table~\ref{table:nextqa_finegrained}. Also, this section includes the comparison with state-of-the-art models of video captioning in Table~\ref{table:captioning}.

\input{tables/activitynet_question_types.tex}
\input{tables/nextqa_val_finegrained.tex}
\input{tables/videocaptioning_sota_comparison.tex}

\section{Qualitative Examples}
\label{appendix:qualitative}

\begin{figure*}[t!] 
\centering
\includegraphics[width=0.8\textwidth]{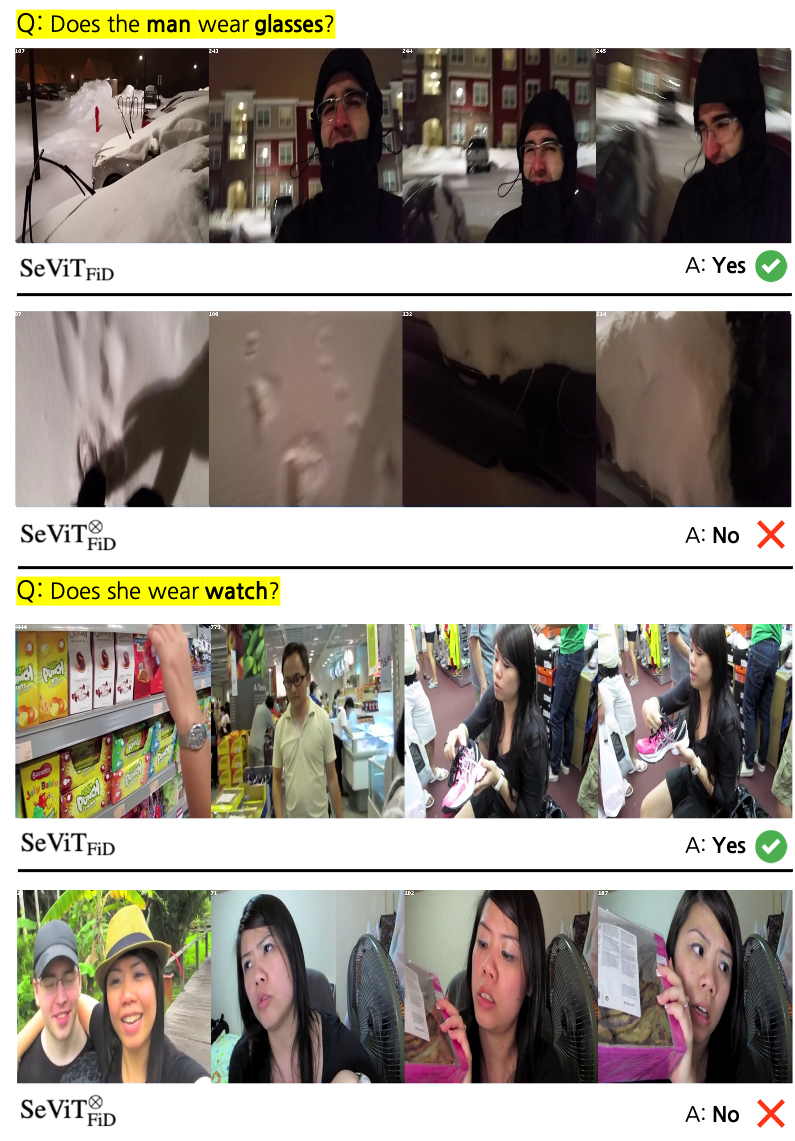}
\caption{We show efficacy of \frameworkname{} by qualitative examples. To this end, we compare \frameworkname$_\text{FiD}$ and \frameworkname$_\text{FiD}^{\otimes}$ in videos \textit{longer than 10 minutes} from Activitynet-QA. The results show that a baseline based on frame sampling, \frameworkname$_\text{FiD}^{\otimes}$, fails to find relevant frames from the long videos while \frameworkname$_\text{FiD}$ successfully finds query-relevant frames.}
\label{fig:qualitative_example}
\end{figure*}

\begin{figure*}[t!] 
\centering
\includegraphics[width=0.8\textwidth]{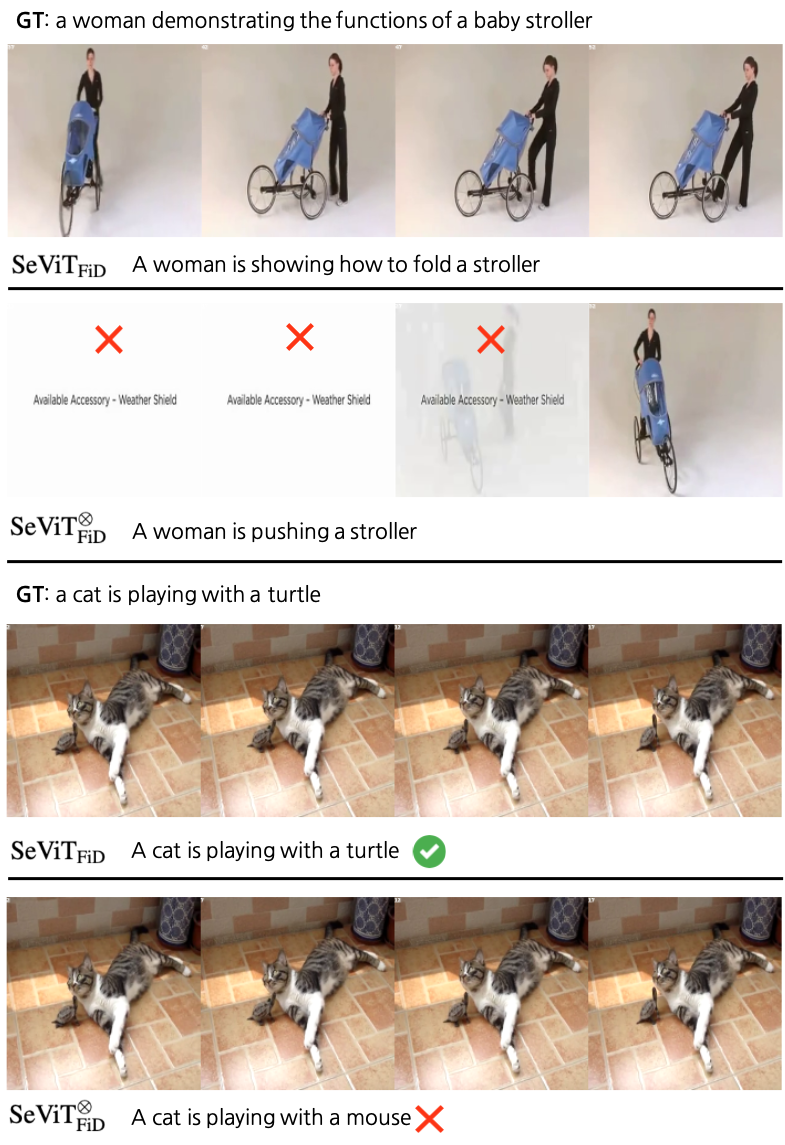}
\caption{Qualitative examples of video captioning. We compare \frameworkname$_\text{FiD}$ and \frameworkname$_\text{FiD}^{\otimes}$ on videos in MSRVTT-Caption (top) and MSVD-Caption (bottom). Especially, the top example shows the case that frames by \frameworkname$_\text{FiD}^{\otimes}$, \ie, random frame sampling, include uninformative frames resulting in performance degradation.}
\label{fig:qualitative_caption}
\end{figure*}

In Figure~\ref{fig:qualitative_example}, we show qualitative examples comparing \frameworkname{}$_\text{FiD}$ and \frameworkname{}$_\text{FiD}^{\otimes}$ on the \textit{10 minute-long videos} in Activitynet-QA. We also include qualitative examples of video captioning in Figure~\ref{fig:qualitative_caption}.

%% file: tables/dataset.tex
\begin{table}[!h]
    \centering
    \small
    \caption{Dataset statistics used in our experiments.}
    \vskip 0.15in
    \begin{adjustbox}{width=0.48\textwidth} 
    \begin{tabular}{lcccccc}
        \toprule
        \multirow{2}[3]{*}{Dataset} & \multicolumn{3}{c}{Video} & \multicolumn{3}{c}{QA pairs} \\
        \cmidrule(r){2-4} \cmidrule(r){5-7}
        & Train & Val & Test & Train & Val & Test \\
        \midrule
        TGIF-QA & 62,841 & - & 9,575 & 139,395 & - & 25,751\\
        MSVD-QA & 1,200 & 250 & 520 & 30,933 & 6,415 & 13,157 \\
        MSRVTT-QA & 6,513 & 497 & 2,990 & 158,581 & 12,278 & 72,821 \\
        iVQA & 5,994 & 2,000 & 2,000 & 5,994 & 2,000 & 2,000 \\
        Next-QA & 3,870 & 570 & 5,440 & 34,132 & 4,996 & 8,564 \\
        Activitynet-QA & 3,200 & 180 & 800 & 32,000 & 18,000 & 8,000 \\
        \midrule
        MSVD-Cap. & 1,200 & 100 & 670 & - & - & - \\
        MSRVTT-Cap. & 6,513 & 497 & 2,990 & - & - & - \\
        \bottomrule
    \end{tabular}
    \end{adjustbox}
    \label{table:dataset}
\end{table}

%% file: tables/hyperparameter.tex
\begin{table}[t!]
    \centering
    \small
    \caption{Common hyper-parameter setups in our experiments.}
    \vskip 0.15in
    \setlength{\tabcolsep}{10pt}
    \begin{tabular}{lc}
        \toprule
        Hyper-parameter & Setup \\
        \midrule
        learning rate (lr) & 3e-5 \\
        \# epoch & 5 \\
        lr scheduling & linear \\
        warmup ratio & 0.01 \\
        weight decay & 0.01 \\
        label smoothing & 0.1 \\
        \bottomrule
    \end{tabular}
    \label{table:common_hyper}
\end{table}

\begin{table}[t!]
    \centering
    \small
    \caption{Task-specific hyper-parameter setups in our experiments.}
    \vskip 0.15in
    \begin{adjustbox}{width=0.48\textwidth} 
    \begin{tabular}{lccc}
        \toprule
        Dataset & Batch Size & Max input len. & Max output len. \\
        \midrule
        TGIF-Action & 16 & 64 & 10 \\
        TGIF-Transition & 16 & 64 & 10 \\
        TGIF-Frame & 16 & 32 & 10 \\
        MSVD-QA & 16 & 32 & 10 \\
        MSRVTT-QA & 32 & 32 & 10 \\
        iVQA & 16 & 32 & 10 \\
        Next-QA & 16 & 64 & 16 \\
        Activitynet-QA & 32 & 32 & 10 \\
        \midrule
        MSVD-Caption & 16 & 10 & 32 \\
        MSRVTT-Caption & 32 & 10 & 32 \\
        \bottomrule
    \end{tabular}
    \end{adjustbox}
    \label{table:specific_hyper}
\end{table}

%% file: tables/activitynet_question_types.tex
\begin{table*}[t!]
    \centering
    \small
    \caption{Breakdown results on Activitynet-QA test set by fine-grained question types.}
    \vskip 0.15in
    \vskip 0.15in
   \begin{adjustbox}{width=0.85\textwidth} 
    \begin{tabular}{lcccccccccc}
        \toprule
        Model & Motion & Spatial & Temporal & Yes/No & Color & Object & Location & Number & Other & All \\
        \midrule
        JustAsk & 28.0 & 17.5 & \textbf{4.9} & 66.3 & 34.3 & 26.7 & 35.8 & 50.2 & 36.8 & 39.0 \\
        MERLOT & \textbf{33.9} & 18.1 & 4.0 & 72.5 & 36.2 & 24.5 & 36.5 & 51.7 & 37.8 & 41.4 \\
        \midrule
        \frameworkname{}$_\text{MAR}$ & 31.8 & 24.6 & 3.5 & \textbf{78.6} & 64.0 & \textbf{36.2} & \textbf{40.9} & 56.9 & 39.1 & 47.2 \\
        \frameworkname{}$_\text{FiD}$ & 30.6 & \textbf{24.8} & 4.4 & \textbf{78.6} & \textbf{65.4} & 32.1 & \textbf{40.9} & \textbf{59.4} & \textbf{40.3} & \textbf{47.6} \\
        \bottomrule
    \end{tabular}
    \end{adjustbox}
    \label{table:activitynet_question}
\end{table*}

%% file: tables/nextqa_val_finegrained.tex
\begin{table*}[t!]
    \centering
    \small
    \caption{Breakdown results on Next-QA validation set by fine-grained question types.}
    \vskip 0.15in
    \vskip 0.15in
   \begin{adjustbox}{width=0.85\textwidth} 
    \begin{tabular}{lccccccccccc}
        \toprule
        \multirow{2}[3]{*}{Model} & \multicolumn{3}{c}{Causal} & \multicolumn{3}{c}{Temporal} & \multicolumn{4}{c}{Descriptive}& \multirow{2}[3]{*}{All} \\
        \cmidrule(r){2-4}
        \cmidrule(r){5-7}
        \cmidrule(r){8-11}
        & Why & How & All & Bef\&Aft & Present & All & Count & Location & Other & All & \\
        \midrule
        HGA & 47.0 & 44.2 & 46.3 & 49.5 & 52.5 & 50.7 & 44.1 & 72.5 & 55.4 & 59.3 & 49.7 \\
        \midrule
        \frameworkname{}$_\text{MAR}$ & 54.2 & \textbf{51.8} & 53.5 & \textbf{51.0} & 58.4 & 54.0 & 55.4 & 81.7 & 65.2 & 69.2 & 56.1 \\
        \frameworkname{}$_\text{FiD}$ & \textbf{55.5} & 49.9 & \textbf{54.0} & 50.2 & \textbf{59.7} & \textbf{54.1} & \textbf{56.5} & \textbf{82.4} & \textbf{69.2} & \textbf{71.3} & \textbf{56.7} \\
        \bottomrule
    \end{tabular}
    \end{adjustbox}
    \label{table:nextqa_finegrained}
\end{table*}

%% file: tables/videocaptioning_sota_comparison.tex
\begin{table*}[t!]
    \centering
    \small
    \caption{Comparison with baseline models including state-of-the-art models on two video captioning datasets. $\dagger$ indicates \frameworkname{} is initialized with OFA-Large~\cite{wang2022unifying} backbone for VGT-generator. BLEU-4, METEOR, ROUGE-L, CIDEr are reported~\cite{papineni2002bleu, banerjee2005meteor, lin2004rouge, vedantam2015cider}. \textbf{Bold} indicates the best score and \underline{underline} indicates the second best score.}
    \vskip 0.15in
    \begin{tabular}{lccccccccc}
        \toprule
        \multirow{2}[3]{*}{Model} & \multirow{2}[3]{*}{\makecell{Video-Text \\ Pre-training}} & \multicolumn{4}{c}{MSRVTT} & \multicolumn{4}{c}{MSVD} \\
        \cmidrule(r){3-6} \cmidrule(r){7-10}
        & & BLEU & METEOR & ROUGE & CIDEr & BLEU & METEOR & ROUGE & CIDEr \\
        \midrule
        DECEMBERT & \cmark & 45.2 & \textbf{39.7} & \underline{64.7} & 52.3 & - & - & - & - \\
        MV-GPT & \cmark & \textbf{48.9} & \underline{38.7} & 64.0 & 60.0 & - & - & - & - \\
        LAVENDER & \cmark & - & - & - & 60.1 & - & - & - & \textbf{150.7} \\
        \midrule
        ORG-TRL & \xmark & 43.6 & 28.8 & 62.1 & 50.9 & 54.3 & 36.4 & 73.9 & 95.2 \\
        SwinBERT & \xmark & 41.9 & 29.9 & 62.1 & 53.8 & 58.2 & 41.3 & 77.5 & 120.6 \\
        \midrule
        \textit{Ours}  & & & & & & & & & \\
        \frameworkname{}$_\text{MAR}^{\otimes}$ & \xmark & 44.9 & 30.7 & 63.0 & 58.6 & 64.0 & 42.9 & 79.5 & 127.4 \\
        \frameworkname{}$_\text{MAR}$ & \xmark & 44.9 & 31.0 & 63.0 & 57.6 & 64.6 & 42.9 & 80.3 & 136.1 \\
        \frameworkname{}$_\text{FiD}^{\otimes}$ & \xmark & 46.2 & 31.4 & 64.3 & 61.8 & 65.7 & 43.3 & 79.8 & 134.9 \\
        \frameworkname{}$_\text{FiD}$ & \xmark & 47.1 & 31.6 & 64.4 & \underline{62.2} & \underline{66.9} & \underline{43.7} & \underline{80.9} & 135.5 \\
        \midrule
        \frameworkname{}$_\text{FiD}^{\dagger}$ & \xmark & \underline{48.2} & 31.7 & \textbf{64.9} & \textbf{63.7} & \textbf{69.1} & \textbf{46.3} & \textbf{83.0} & \underline{148.1} \\
        \bottomrule
    \end{tabular}
    
    \label{table:captioning}
\end{table*}